\documentclass{article}
\usepackage{ijcai19}

\usepackage{times}
\usepackage{soul}
\usepackage{url}
\usepackage[small]{caption}
\usepackage{graphicx}
\usepackage{amsmath}
\usepackage{booktabs}
\usepackage{algorithm}
\usepackage{algorithmic}
\usepackage{bchart}
\usetikzlibrary{fit}
\usepackage{pgfplots}
\usepackage[makeroom]{cancel}

\title{ %Alpha-blending: a novel neural network quantization training flow  
Learning low-precision neural networks without\\ Straight-Through Estimator (STE) }

\iftrue
\author{
Zhi-Gang Liu
\and
Matthew Mattina
\affiliations
Arm Machine Learning Research Lab.
\emails
\{zhi-gang.liu, matthew.mattina\}@arm.com
}
\fi

\begin{document}

\maketitle

\begin{abstract}

The Straight-Through Estimator $(\textbf{STE})$ \cite{ste:}\cite{Bengio2013STE} is widely used for back-propagating gradients through the quantization function, but the STE technique lacks a complete theoretical understanding. We propose an alternative methodology called alpha-blending $(\textbf{AB})$, which quantizes neural networks to low precision using stochastic gradient descent $(\textbf{SGD})$. Our $(\textbf{AB})$ method avoids STE approximation by replacing the quantized weight in the loss function by an affine combination of the quantized weight $\textbf{w}_q$ and the corresponding full-precision weight $\textbf{w}$ with non-trainable scalar coefficient $\alpha$ and $(1- \alpha)$. During training, $\alpha$ is gradually increased from 0 to 1; the gradient updates to the weights are through the full precision term, $(1-\alpha)\textbf{w}$, of the affine combination; the model is converted from full-precision to low precision progressively. To evaluate the $(\textbf{AB})$ method, a 1-bit BinaryNet \cite{binary} on CIFAR10 dataset and 8-bits, 4-bits MobileNet v1, ResNet\_50 v1/2 on ImageNet are trained using the alpha-blending approach, and the evaluation indicates that $\textbf{AB}$ improves top-1 accuracy by 0.9\%, 0.82\% and 2.93\% respectively compared to the results of STE based quantization \cite{binary}\cite{TF-mb-int8:}\cite{TF-rn-int8:}\cite{TF-int4-8:}.     
\end{abstract}

\section{Introduction}

Deep Neural Networks (DNNs) have demonstrated outstanding performance on a wide range of tasks, including image classification ~\cite{image-classification}, speech recognition ~\cite{speech} etc. These networks typically consists of multiple convolution layers with a large number of parameters. The models are trained on high performance servers typically with GPUs and are deployed on lower-end machines, i.e. mobile or IoT devices, for inference tasks. Improved inference accuracy usually comes with  millions of model parameters and high computation cost. For example, the largest Mobilenet v1 model ~\cite{mobilenet} has 4.2 million parameters and 569 million floating point MAC per inference ~\cite{TF-mb-int8:}. For applications that demand high inference accuracy, low latency and low power consumption, the large memory requirements and computation costs are a significant challenge for constrained platforms.
\parskip = \baselineskip

To achieve efficient inference, one approach is to design compact network architectures from scratch  ~\cite{compact-network-1}~\cite{compact-network-2}~\cite{compact-network-3}~\cite{compact-network-4}.  Alternatively, existing models can be optimized for efficiency. There are several optimization techniques that boost efficiency when applied to pretrained models: weight pruning ~\cite{pruning-cluster}~\cite{pruning-svd-cluster}, weight clustering ~\cite{pruning-cluster}~\cite{pruning-svd-cluster}, singular value decomposition (SVD) ~\cite{svd} and quantization ~\cite{50}~\cite{quant-2}~\cite{quant-3}~\cite{quant-4:}. The basic principle is to reduce the number of parameters and/or lower the computation cost of inference. Weight pruning techniques remove parameters while minimizing the impact on inference accuracy. Weight clustering clusters similar weights to shrink the overall size of a model.  The $\textit{SVD}$ method potentially reduces both model size and computation cost through discarding small singular values. Quantization techniques convert normal floating-point values to narrow and cheaper integer or fixed point i.e. 8-bits, 4-bits or binary multiplication operations without incurring significant loss in the accuracy. There are three major benefits to quantization: reduced memory bandwidth, reduced memory storage, and higher throughput computation. The predominant numerical format used for training neural networks is IEEE fp32 format. There is a potential 4x reduction in overall bandwidth and storage if one can quantize fp32 floating point to 8-bits for both weight and activation. The corresponding energy and area saving are 18x and 27x ~\cite{dally} respectively. The efficient computation kernel libraries for fast inference, i.e. Arm CMSIS ~\cite{arm-cmsis}, Gemmlowp ~\cite{gemmlowp}, Intel MKL-DNN ~\cite{mkl}, Nvidia TensorRT ~\cite{nvidia} and custom ASIC hardware, are built upon the reduced precision numerical forms. 

\parskip = \baselineskip

The Straight-Through Estimator (STE) \cite{ste:}\cite{Bengio2013STE} is widely implemented in discrete optimization using SGD due to its effectiveness and simplicity. STE is an empirical workaround to the gradient vanishing issue in Backprop; however it lacks complete mathematical justification especially for large-scale optimization problems \cite{Yin2018STE}. In this paper, we propose a novel optimization technique, termed alpha-blending ($\textbf{AB}$), for quantizing full precision networks to lower precision representations(8-bits, 4-bits or 1-bit). AB does not rely on the concept of STE to back-propagate the gradient update to weights; AB instead replaces the weight vector $\textbf{ w }$ in the loss function by the expression $\textbf{w}_{ab} = (1 - \alpha) \textbf{w} + \alpha  \textbf{w}_q $, which is the affine combination of the  $\textbf{w}$ and its quantization $\textbf{w}_q $. During training, we gradually increase the non-trainable parameter $\alpha$ from 0.0 to 1.0. This formulation isolates the quantized weights $\textbf{w}_q$  from the full-precision trainable weights $\textbf{w}$ and therefore avoids the challenges arising from the use of Straight-Through Estimation (STE).

\parskip = \baselineskip

To evaluate the performance of the proposed method, we trained single-bit BinaryNet \cite{binary} on CIFAR10 and 4-bits, 8-bits MobileNet v1, ResNet v1 and v2 models on the ImageNet dataset. $\textbf{AB}$ outperforms previous state-of-art STE based quantization 0.9\% for 1-bit BinaryNet and 2.9\% for 4-bits weight and 8-bits activation (4-8) ~\cite{TF-int4-8:} in top-1 accuracy. Moreover, we have applied our AB approach to quantize MobileNet v1, ResNet v1,2 networks with both 4-bit weight as well as 4-bit activation (4b/4b). In this configuration, our 4b/4b quantization delivers similar accuracy level as the best known 4b/8b quantization approach \cite{TF-int4-8:}. 

\section{Related works}

There is a significant body of research on neural network quantization techniques from the deep learning community. BinaryConnect ~\cite{50} binarizes the weights of neural networks using the sign function. Binary Weight Network ~\cite{42} has the same binarization while introducing a scaling factor. BinaryNet ~\cite{52}~\cite{binary} quantizes both weights and activations to binary values. TWN ~\cite{compact-network-4} constructs networks with ternary values 0, +/-1 to balance the accuracy and model compression compared to the binary quantization.
STE \cite{ste:} is used to approximate the gradient of quantization function during the learning process. Once they are quantized, these models eliminate majority of the floating-point multiplications, and therefore exhibit improved power efficiency by using SIMD instructions on commodity micro-processor or via special hardware. On the downside, the single bit quantization schemes often lead to substantial accuracy drop on large scale dataset while achieving good results on simple dataset such as MNIST, CIFAR10.

\parskip = \baselineskip

 Another approach is to train the network in full floating-point domain, then statically quantize the model parameter into reduced numerical forms and keep the activation in floating-point. Google’s Tensorflow provides a post-training quantization flow ~\cite{TF-quant:} to convert float-point weights into 8-bits of precision from – INT8. Its uniform affine quantization maps a set of floating-point values to 8-bits unsigned integers by shifting and scaling ~\cite{TF-int4-8:}. The minimum and maximum values correspond to quantized value 0 and 255 respectively. Another mapping scheme is uniform symmetric quantizer, which scales the maximum magnitude of floating-point values to maximum 8-bit integer e.g. 127 and the floating-point zero always mapped to quantized zero. The conversion is done once, and reduction of model size is up to 4X. A further improvement dynamically quantizes activations into 8-bits as well at inference. With 8-bits weight and activation, one can switch the most compute-intensive operations e.g. convolution, matrix multiply (GEMM) from original floating-point format to the cheaper operation, and reduces the latency as well. 

\parskip = \baselineskip

 The main drawback of such post-processing approach is the degradation in model accuracy. To overcome this accuracy drop, “quantization aware training” ~\cite{TF-quant:} techniques have been developed to ensure that the forward pass uses the reduced precision for both training and inference.  To achieve this, full precision weights and activations values flow through “fake quantization” nodes, then quantized values feed through convolution or matrix multiply. Applying the Straight-Through Estimator (STE) approximation ~\cite{ste:}~\cite{binary}~\cite{arXiv180805240Y}, the operations in the back propagation phase are still at full precision  as this is required to provide sufficient precision in accumulating small adjustment to the parameters.

\section{Alpha-blending, the proposed method ($\textbf{AB}$)}

We introduce an optimization methodology, alpha-blending ($\textbf{AB}$), for quantizing neural networks. Section \ref{sect:3.1} describes the scheme of $\textbf{AB}$ and weights quantization; section \ref{sect:3.2} sketches the quantization of activation using $\textbf{AB}$.        

\subsection{Alpha-blending \textbf{AB} and quantization of weights} \label{sect:3.1}
\begin{figure}[h!] 
\centering
\includegraphics[scale=0.6]{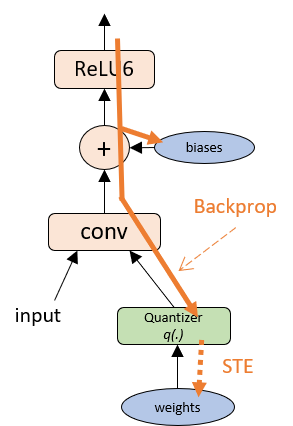}
\caption{ Gradient update to the full precision weight in backprop using STE approximation as eq. \ref{eq:2'} and \ref{eq:2}.} 
\label{fig:bp_ste}
\end{figure}

During quantization-aware training, the full precision weights are quantized to low precision values ${ { \textbf{w}_q } }$. Mathematically, we want to minimize a convex function $L(\textbf{w})$ as equation \ref{eq:1} with the additional constraint that $\textbf{w}$ must be n-bit signed integers i.e. $\textbf{w} \in \textbf{Q} = [-(2^{n-1}-1), 2^{n-1}-1]$.
\begin{equation} \label{eq:1}
        \min\limits_{s.t.\ w \in \textbf{Q}} L(\textbf{w})
\end{equation}
 Previous approaches i.e. ~\cite{TF-quant:},~\cite{binary} insert quantizer nodes in the computation graph. These nodes receive full precision input $\textbf{w}$ and generate quantized output $\textbf{w}_q = \textit{q}(\textbf{w})$, between the full precision weights $\textbf{w}$ and computation nodes as in Figure \ref{fig:bp_ste}. The quantized weights $ \textbf{w}_q = \textit{q}(\textbf{w}) $ are used in the forward and backward pass while the gradient update to the full precision weight uses full precision to ensure smooth updates to the weights. But the quantization function has zero gradient almost everywhere $ {\partial{\textbf{w}_q}}/{\partial{\textbf{w}}} \underset{a.e.}{=} 0 $, which prevents further backpropagation of gradients and halts learning. The $\textit{Straight-Through Estimator}$ (STE) ~\cite{ste:}~\cite{binary}~\cite{TF-int4-8:} was developed to avoid the vanishing gradient problem illustrated in Figure \ref{fig:bp_ste}. \textbf{STE} approximates quantization with the identity function $ \textit{I}(\textbf{w}) = \textbf{w}$ in $\textbf{Backprop}$ as eq. \ref{eq:2'}. Therefore with STE, the gradient of the quantization function with respect to the full precision weight is approximated using the quantized weight as in equation \ref{eq:2}. We hypothesize that the error introduced by this approximation may impact the accuracy of the gradient computation, thereby degrading overall network accuracy, especially for very low precision (1-bit or 4-bit) networks.
%\begin{equation} \label{eq:5}
%L(\textbf{w}) = \| \textbf{a}_L - \textbf{y} \|^2  +  %R(\textbf{w}_k)
%\end{equation}
%\begin{equation} \label{eq: 6}
%{\textbf{a}}_{k+1} = \delta(\textbf{w}_k \textbf{a}_k + %{\textbf{b}}_k ) \ \ \ k\in \big[1,L-1\big] 
%\end{equation}
\begin{equation} \label{eq:2'}
\frac{\partial{\textbf{w}_q}}{\partial{\textbf{w}}} = \frac{\partial{\textit{q}(\textbf{w})}}{\partial{\textbf{w}}} \underset{\scriptsize{STE}}{ \approx } \frac{\partial{\textit{I}(\textbf{w})}}{\partial{\textbf{w}}} = 1
\end{equation}
\begin{equation} \label{eq:2}
    \frac{\partial{L(\textbf{w})}}{\partial{\textbf{w}}} =
    \frac{\partial{L(\textbf{w}_q)}}{\partial{\textbf{w}_q}} \cdot \frac{\partial{\textbf{w}_q}}{\partial{\textbf{w}}} \underset{STE}{\approx}  \frac{\partial{L(\textbf{w}_q)}}{\partial{\textbf{w}_q}}
\end{equation}

\begin{figure}[h!] 
\centering
\includegraphics[scale=0.6]{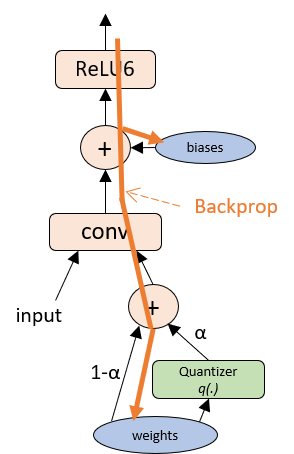}
\caption{\textbf{AB} quantization performs the convolution using an affine combination of the full precision weights and the quantized weights. The coefficient $\alpha$ is gradually increased from 0 to 1 during training. This approach avoids  back-propagation through the Quantizer, eliminating the gradient vanishing path (from the quantized weight node in light green to the weight node in blue). There is no need to apply $\textit{Straight Trough Estimator}$ (STE) during the $\textit{Backprop}$. The actual weight gradient update goes through the $(1-\alpha)$ path, where the gradient, eq. \ref{eq:12}, is well-defined.}  
\label{fig:bp_no_ste}
\end{figure}

\begin{figure}
\resizebox {\columnwidth} {!} {
%\begin{minipage}
\begin{tikzpicture}

%\pgfplotsset{%scale only axis,}
\begin{axis}[
    axis y line*=left,
    title={Mobilenet\_v1\_0.25/128, top-1 accuracy with weight $\textbf{w}$ and quantized weight $\textbf{q}$ },
    xlabel={training step},
    ylabel={Accuracy},
    xmin=0, xmax=100000,
    ymin=0.2, ymax=0.5,
    xtick={0,20000,40000,60000,80000,100000},
    ytick={0.20,0.30,0.40,0.5},
    legend pos=north west,
    ymajorgrids=true,
    grid style=dashed,
]
 
\addplot[
    %xscale=0.8,
    color=blue,
    mark=o,
    ]
    coordinates {
    (0,0.415)(234,0.41)(3770,0.412)(7306,0.41)(10995,0.407)(17445,0.398)(20907,0.39)(24375,0.386)(31320,0.371)(34818,0.363)(41830,0.349)(48828,0.34)(55746,0.329)(66537,0.318)(77303,0.313)(94916,0.308)(98354,0.307)
    }; \label{fig:3_curves_weight}
    
    \addlegendentry{weight $ \textbf{ w } $}
    
 \addplot[
    %xscale = 0.8,
    color=green,
    mark=triangle,
    ]
    coordinates {
    (0,0)(234,0.27)(3770,0.289)(7306,0.305)(10995,0.319)(17445,0.346)(20907,0.357)(24375,0.368)(34818,0.389)(41830,0.398)(48828,0.404)(55746,0.406)(66537,0.408)(77303,0.407)(94916,0.41)(98354,0.41)
    };\label{fig:3_curves_quantized}
    \addlegendentry{quantized weight $\textbf{w}_q $}
\end{axis}

 \begin{axis}[
    axis y line*=right,
    hide x axis,
    ylabel={ $\alpha \  (\textbf{alpha}$)},
    ylabel near ticks,
    xmin=0, xmax=100000,    
    ymin=0.0, ymax=1.1,
    legend pos=south east,
    %ymajorgrids=true,
    %grid style=dashed,
]
\addplot[
    color=red,
    mark=none,
    ]
    coordinates {
    (0,0)(234,0.007)(3770,0.109)(7306,0.20355)(10995,0.2949)(17445,0.4374)(20907,0.505)(24375,0.567)(31320,0.676)(34818,0.723)(41830,0.803)(48828,0.866)(55746,0.913)(66537,0.9625)(77303,0.988)(94916,0.9998)(98354,1.0)(100000,1.0)
    };\label{fig:3_curves_alpha}
    \addlegendentry{$ \alpha=\textit{A}(step)$}
\end{axis}
\end{tikzpicture}
%\end{minipage}
}
\captionof{figure}{Two accuracy curves, evaluated with the full precision weights $\textbf{w}$ and 8-bits quantized weights $\textbf{w}_q $ during $\textbf{AB}$ quantization training with Mobilenet 0.25/128 V1 for 2.5 epochs. The $ \alpha $ curve \ref{fig:3_curves_alpha} is the $ \textbf{A} $ function in eq. \ref{eq:13}. The accuracy \ref{fig:3_curves_weight} corresponding to full precision weights has dropped 10\% during the training while the \ref{fig:3_curves_quantized}, with the quantized weights, has gradually increased to approach its maximum accuracy 40.9\% when $\alpha$ = 1.0. The final quantized model has 0.6\% accuracy loss compared to full precision one.
}      
\label{fig:3_curves}
\end{figure}
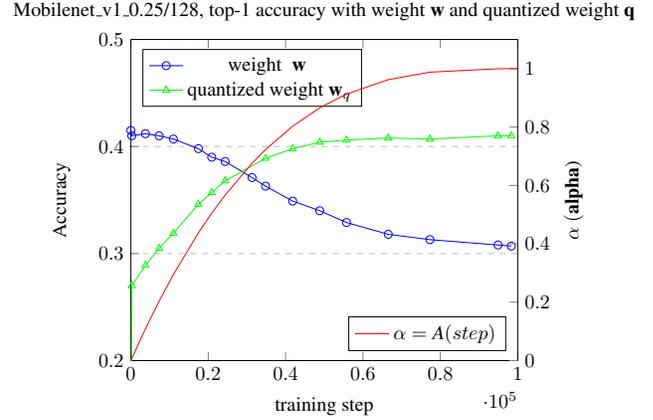

Our proposed method, alpha-blending (\textbf{AB}), does not rely on the $\textit{Straight-Through Estimator}$ (STE) to overcome the quantizer's vanishing gradient problem in $\textbf{Backprop}$, therefore it eliminates the quantization error due to equation \ref{eq:2}. \textbf{AB} replaces the weight term in the loss function by $ (1-\alpha)\textbf{w} + \alpha \textbf{w}_q $, an affine combination of the original full precision weight term and its quantized version with coefficient $ \alpha $. The new loss function $L_{ab}(\textbf{w},\alpha)$ for a neural network is shown in equation \ref{eq:10}. The gradient of $L_{ab}(\textbf{w}, \alpha)$ with respect to the weights is in equation \ref{eq:12}, accepting the $\textbf{zero}$ gradient of quantization function  $ {\partial{\textbf{w}_q}}/{\partial{\textbf{w}}} \underset{a.e.}{=} 0 $ without STE approximation. Its Backprop flow is illustrated in figure \ref{fig:bp_no_ste}.     
\begin{equation} \label{eq:10}
\begin{aligned}
L_{ab}(\textbf{w}, \alpha) = L((1-\alpha)\textbf{w} + \alpha \textbf{w}_q)
\end{aligned}
\end{equation}
\begin{equation} \label{eq:12}
\begin{aligned}
    \frac{\partial{L_{ab}}}{\partial{\textbf{w}}} = (1-\alpha+ \underset{=0\ a.e.}{\xcancel{\alpha \frac{\partial{\textbf{w}_q}}{\partial{\textbf{w}}}}}) 
    \left.\frac{\partial{L(\textbf{w}^{'})}}{\partial{\textbf{w}^{'}}} \right|_{\textbf{w}^{'}= (1-\alpha)\textbf{w} +\alpha\textbf{w}_q}
\end{aligned}
\end{equation}
 The $\textbf{AB}$ flow gradually increases the non-trainable parameter $ \alpha $ from 0 to 1 using a function of the form shown in equation \ref{eq:13} for training steps in the optimization window $[T_0, T_1]$. An example is shown in Figure \ref{fig:3_curves}. The function in equation \ref{eq:13} is not unique, for example, an alternative choice is $ A(step, \lambda) = 1 - e^{- \lambda \cdot step} $. The optimization window $[T_0, T_1]$, during which $\alpha$ is increased, is a user-defined hyper parameter. 

We use algorithm \ref{alg:IPQ} described in section \ref{sect:exp-ppq} to convert $\textbf{w}$ to $\textbf{w}_q = {\gamma}_w \cdot \textbf{q}_w $, where $ {\gamma}_w$ is a scaling factor and $\textbf{q}_w \in \textbf{Q}$, at certain frequency, $\textit{quantizing\_frequency}$, in training steps.    
\begin{equation} \label{eq:13}
\begin{aligned}
A(step) = 
\begin{cases}
    0  & step \le T_0 \\
    1 - (\frac{T_1 - step}{T_1-T_0})^3 & T_0 < step \le T_1  \\
    1 & T_1 < step   
\end{cases}
\end{aligned}
\end{equation}
\begin{algorithm}[tb]
   \caption{Alpha-blending optimization ($\textit{ABO}$)}
   \label{alg:ABO}
\begin{algorithmic}
    \STATE {\bfseries Input:} $\textit{derivative loss function}     \ L(\textbf{w})$ 
    \STATE {\bfseries Def.} function: $L_{ab}(\textbf{w}, \textbf{w}_q,  \alpha)$ = $ L((1-\alpha)\textbf{w} + \alpha \textbf{w}_q) $ 
    \STATE {\bfseries Initialize:} {$ \textbf{w} \gets w_0, \alpha \gets 0, \varepsilon \gets {learning\_rate}, \textit{f} \gets {optimization\_frequency}, \textit{T}_0,\textit{T}_1 \gets {traing\_window} $}
    \FOR{$step=0$ {\bfseries to} $\textit{T}\ \ $}
    \STATE $ w_q \gets $ Algorithm \ref{alg:IPQ}  ${\textit{PPQ}}(\textbf{w})$ or other optimization function $(\textbf{w})$ 
    \STATE $\textbf{w} \gets \textbf{w} - {\left.  \varepsilon \cdot (1-\alpha)\ \frac{\partial{L(\textbf{w}^{'})       }}{\partial{\textbf{w}^{'}}}\right|_{\textbf{w}^{'} = (1-\alpha)\textbf{w} + \alpha \textbf{w}_q}} $ 
    \IF {$step \ \% \ \textit{f} = 0 $ and $ \alpha < 1 $}
    \STATE $\alpha \gets A(step) \ \ \ \ \ $   
     \COMMENT{Raising $\alpha$ toward to 1.0; eq \ref{eq:13}}
    \ENDIF
   \ENDFOR
   \STATE {\bfseries Output:} $ \textbf{w}_q $
\end{algorithmic}
\end{algorithm}
 Algorithm \ref{alg:ABO} summarizes the $\textbf{AB}$ optimization procedure, in which the original learning rate $\varepsilon$ is scaled by the factor $(1-\alpha)$ to act as an effective learning rate  $\varepsilon \cdot (1-\alpha)$. 
 
\begin{figure}[h!] 
\centering
\includegraphics[scale=0.45]{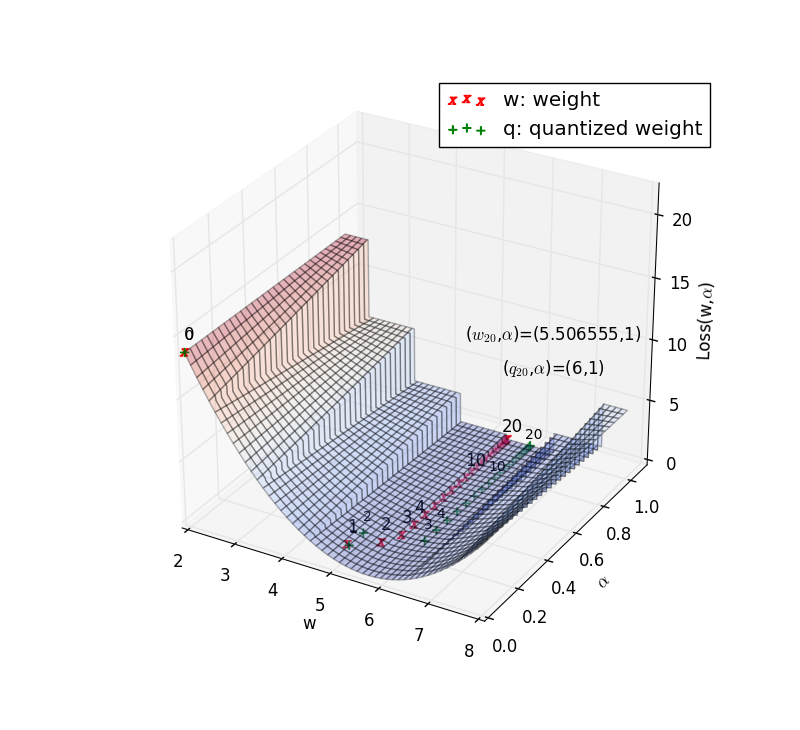}
\caption{Apply \textbf{AB} to minimize a trivial example $ Loss(w) = (w - 5.7)^2 $, equivalently to find the minimal  of 2D surface   
$ Loss(w, \alpha) = ((1 - \alpha)w + \alpha w_q)^2 $ using \textbf{SGD} while alpha $(\alpha)$ has changed from 0 to 1 using function ${\textit{A}}(\cdot)$ in eq. \ref{eq:13}, and $w_q=round(w)$. ${w}$ started at the initial value $(w,\alpha)$=$(2.0,0)$ and moved along the ${\color{red}\small{X}}$ trace, its corresponding quantized weights are marked by $\color{green}+$. In 20 steps, the iteration converged to ($\textbf{w}$,$\alpha$)=(5.506555, 1). ${w}_q$ = 6 is the final quantized solution.}  
\label{fig:AB_3d}
\end{figure}

 To visualize the process, figure \ref{fig:AB_3d} demonstrates how to solve the trivial example, $\underset{s.t. \ w \in \textbf{Q}}{arg \ min} {(w - 5.7)^2}  = 6 $ using AB.

To compare the $\textbf{AB}$ optimization concept with $\textbf{STE}$, we trained the single bit 8-layer BinaryNet defined in \cite{binary} on the CIFAR10 dataset in section \ref{sect:ab_ste}, figure \ref{fig:alpha-ste}. The top-1 accuracy score achieved with $\textbf{AB}$ is 0.9\% higher compared to the accuracy achieved with $\textbf{STE}$.  

Figure \ref{fig:3_curves} shows a more practical example of $\textbf{AB}$ quantization using MobileNet\_1.0\_0.25/128 v1 on the ImageNet dataset. The $\textbf{AB}$ quantization flow gradually transforms the full precision model at $\alpha=0$ to a model with quantized weights $\textbf{w}_q$ at $\alpha=1.0$ with an accuracy loss of 0.6\% versus the full precision model.      
%\begin{tikzpicture}
%    \begin{axis}[
%        title = {},
%        xmin=2, xmax=8,
%        ymin=0, ymax=1,
%        view={0}{90},
%    ]
%    \addplot3[
%        contour gnuplot = {levels={2.56, 1.28, 0.64, 0.32, 0.16, 0.08, 0.06, 0.04, 0.02, 0.01, 0.005, 0.0}, contour label style={draw=black},},
%        samples=10,
%        contour/draw color={black},
%        ]
%    {((1-y)*x + y*round(x) - 5.7)^2};
%    \end{axis}

\subsection{Quantization of activation} \label{sect:3.2}

$\textbf{AB}$ uses the $\textit{PPQ}$, algorithm \ref{alg:IPQ} in section 4.2, to quantize the input feature maps or activation ${\textbf{a}}$ to ${\textbf{a}}_q$ as well, and accumulates the scaling factor $ {\gamma}_a $ via exponential moving average with the smoothing parameter being close to 1, e.g. 0.99. Thus $\textbf{a}$ can be approximated as $
\textbf{a} \approx {\gamma}_{a} \cdot {\textbf{q}}_a $

For inference ($\alpha$=1), the floating point computation of the $k^{th}$ layer in forward pass is $\textbf{a}^{(k+1)}$ = $\delta({{\textbf{w}}}^{(k)}{\textbf{a}}^{(k)} + {\textbf{b}}^{(k)})$. With the quantizaton of both weight and activation, the same calculation becomes eq. \ref{eq:51}.   
\begin{equation} \label{eq:51}
\begin{split}
      \delta(\textbf{w}\cdot{\textbf{a}}+{\textbf{b}}) \approx 
      \delta({\gamma}_{w} \textbf{q}_w \cdot {{\gamma}_a} {\textbf{q}}_a + {\textbf{b}})  \\ =
      \delta((\gamma_w {{\gamma}_a})(\textbf{q}_w \cdot {\textbf{q}}_a) + {\textbf{b}})
\end{split}
\end{equation}
$(\textbf{q}_w \cdot {\textbf{q}}_a) $ in \ref{eq:51} is the compute-intensive operation of matrix multiply or convolution $(\textbf{GEMM})$ in low precision quantized values, which will gain significant power efficiency compared to the original floating-point version. Other relatively unimportant terms in \ref{eq:51} e.g. $(\gamma_w\gamma_a)$ and $\textbf{b}$ can be represented by higher precision fixed points.       

\section{Experiments}

To evaluate the AB quantization methodology, we performed several experiments. The first one, in section \ref{sect:ab_ste}, is a single bit (1-bit) control test between $\textbf{STE}$ and $\textbf{AB}$ on CIFAR10. Section \ref{sect:exp4-8} presents results for Mobilenet v1 and ResNet v1,2 with the ImageNet ILSVRC 2012 dataset. All evaluations were performed on a x86\_64 ubuntu Linux based Xeon server, Lenovo P710, with a TitanV GPU.

\subsection{BinaryNet with alpha-blending \textbf{AB} and Straight-Through Estimator ($\textbf{STE}$)} \label{sect:ab_ste}

\begin{figure}[h!] 
\centering
\includegraphics[scale=0.40]{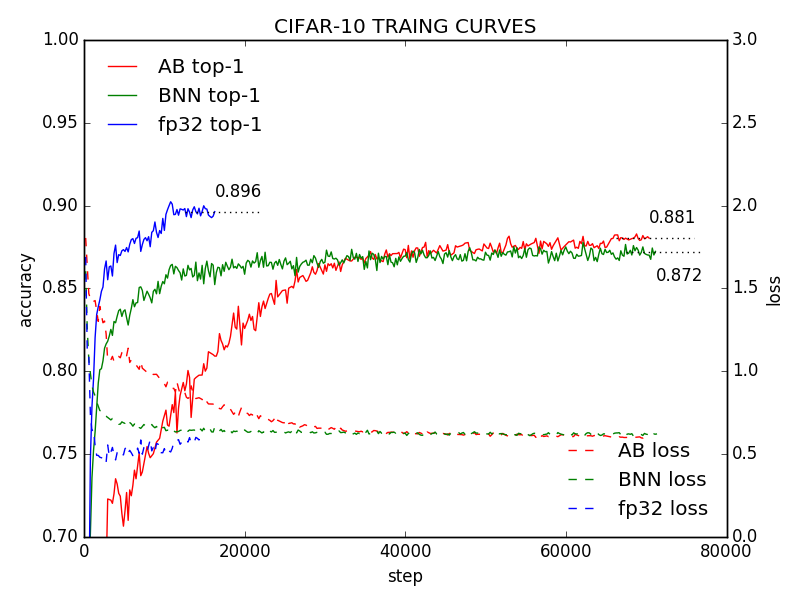}
\caption{Training curves of BinaryNet on CIFAR-10 dataset. The dashed lines represent the validation Loss and the continuous lines are the corresponding validation accuracy. The blue curve of fp32 baseline has max top-1 accuracy 0.896. BNN which utilized $\textbf{STE}$ in training, blue line, converges to 0.872, while the red line of $\textbf{AB}$ yields a better top-1 accuracy 0.881.}      
\label{fig:alpha-ste}
\end{figure}

To evaluate AB's function directly, 1-bit BinaryNet (BNN) \footnote{https://github.com/itayhubara/BinaryNet.tf} \cite{binary} on CIFAR-10 was trained on Tensorflow using AB and STE respectively. Both weight and activation are quantized into +1 or -1 (single bit) by the same binarization function, $ binarize(x) = Sign(x) $. Figure \ref{fig:alpha-ste} shows the results of these experiments. The AB method achieves a top-1 accuracy of 88.1\%. Using STE, we achieve 87.2\%. The FP32 baseline accuracy is 89.6\%.

\subsection{4-bits and 8-bits quantization with MobileNet and ResNet} \label{sect:exp4-8}

In this section, we describe the iterative quantization scheme we use to quantize FP32 values to low-precision, Progressive Projection Quantization ($\textbf{PPQ}$). We apply PPQ to convert floating point values into 4-bits or 8-bits integers, then utilize $\textbf{PPQ}$ and $\textbf{AB}$ to quantize MobileNet and ResNet into 4-bits or 8-bits and compare the result with existing results. All results are consolidated into Figure \ref{fig:cha_1} for easy comparison.    

\subsubsection{Progressive projection quantization, $\textbf{\emph{PPQ}}$} \label{sect:exp-ppq}
To quantize a set of N floating-point values $ \textbf{x} = \big\{ x_i | i \in \big[0,N-1\big] \big\} $ to symmetric n-bits signed integer set $ \textbf{x}_q = \big\{ q_i\ |\ q_i\in{ \textbf{Q}=\big(0,\pm1,\pm2,...,\pm\big(2^{n-1}-1\big)} \big), i \in \big[0,N-1\big]\big\} $ with a positive scaling factor $\gamma$, we can approximate the initial quantization by rounding $ \left.{\frac{x_i}{\gamma}}\right|_{\gamma = \frac{max|x|}{2^{n-1}-1}} $ to the nearest-neighbor in $ \textbf{Q} $ as equation \ref{eq:40}. Then we can improve $ \gamma $ by equation \ref{eq:41}.   
\begin{equation} \label{eq:40}
\begin{aligned}
\textbf{x}_q = round(\frac{\textbf{x}}{\gamma}) 
\end{aligned}
\end{equation}
\begin{equation} \label{eq:41}
\begin{aligned}
\gamma = \frac{\langle  \textbf{x} ,\textbf{x}_q\rangle}{\langle \textbf{x}_q,\textbf{x}_q\rangle}
\end{aligned}
\end{equation}
\parskip = \baselineskip
$\textit{PPQ}$ is an iterative procedure: by repeatedly applying eq. \ref{eq:40} and \ref{eq:41}, as described in algorithm \ref{alg:IPQ}, projects vector $ \textbf{x} $ onto space $ \textbf{Q} $ to determine $ \gamma $ progressively \cite{admm}. 
 The procedure is guaranteed to converge to a local minimum. In practice, convergence is very fast and 3 iterations is enough. Thus, $\textbf{x}$ can be approximated by the product of the scalar $\gamma$ and $\textbf{x}_q $:  $
\textbf{x} \approx \gamma \textbf{x}_q = \gamma \cdot round(\frac{\textbf{x}}{\gamma}) $

\begin{algorithm}[tb]
   \caption{Progressive Project Quantization ($\textit{PPQ}$)}
   \label{alg:IPQ}
\begin{algorithmic}
   \STATE {\bfseries Input:} full precision vector $ \textbf{x} = \big\{ x_i | i \in \big[0,N-1\big] \big\}$, scaling factor $ \gamma $
   \IF{$ \gamma \leq 0$}
   \STATE Initialize $ \gamma \gets \frac{max\big( 
   |\textbf{x}| \big)}{2^{n -  1}-1} $
   \ENDIF
   \REPEAT
   \STATE $ {\gamma}_0 \gets \gamma $ 
   \FOR{$i=0$ {\bfseries to} $N-1$}
   \STATE $ q_i \gets round(\frac{x_i}{{\gamma}_0}) $
   \ENDFOR
   \STATE $ \gamma \gets \frac{\langle  \textbf{x} ,\textbf{q}\rangle}{\langle \textbf{q},\textbf{q}\rangle} $
   \UNTIL{$\gamma$ = ${\gamma}_0$}
   \STATE  {\bfseries Output:} ${\textbf{q}, \gamma }$
\end{algorithmic}
\end{algorithm}

\subsubsection{Evaluation of 8-bits weight and activation (INT8-8)} \label{sect:4.2}

The top-1 accuracy for 8-bit weight and 8-bit activation quantization are listed in table 1. The $2^{nd}$ column gives the fp32 accuracy of the pre-trained models ~\cite{TF-mb-int8:}. The $3^{rd}$ column contains the quantization results \cite{TF-mb-int8:}~\cite{TF-rn-int8:}. The last column gives the best results that $\textbf{AB}$ generated. 

 Both quantization approaches delivered roughly the same top-1 accuracy, although AB has slightly (0.82\%) better accuracy on average. 

\begin{table}[h]
\centering
\caption{top-1 accuracy of fp32 pre-trained models, Tensorflow's INT8-8 and $\textbf{AB} $ 8-8. $^*$ \protect\cite{TF-int4-8:}}
\label{table1}
\begin{tabular}{|l|c|c|c|}
\hline
Model name        & \multicolumn{1}{l|}{

{\begin{tabular}[c]{@{}l@{}}
{fp32} \\ {\ \%} \end{tabular}}

} & \multicolumn{1}{l|}{

{\begin{tabular}[c]{@{}l@{}}
{TF8-8} \\ {\ \ \ \%} \end{tabular}}

} & \multicolumn{1}{l|}{

{\begin{tabular}[c]{@{}l@{}}
{AB8-8} \\ {\ \ \ \ \%} \end{tabular}}

} \\ \hline
MB\_1.0\_224v1  & 70.9                    & 70.1                         & 70.9                         \\ \hline
MB\_1.0\_128v1  & 65.2                    & 63.4                         & 65.0                         \\ \hline
\small{MB\_0.75\_224v1} & 68.4                    & $67.9$                      & 68.2                         \\ \hline
\small{MB\_0.75\_128v1} & 62.1                    & $59.8$                      & 61.6                         \\ \hline
MB\_0.5\_224v1  & 63.3                    & $62.2$                      & 63.0                         \\ \hline
MB\_0.5\_128v1  & 56.3                    & 54.5                         & 55.8                         \\ \hline
\small{MB\_0.25\_224v1} & 49.8                    & 48                           & 49.2                         \\ \hline
\small{MB\_0.25\_128v1} & 41.5                    & 39.5                         & 40.9                         \\ \hline
ResNet\_50v1    & 75.2                    & $75^{\small{*}}$                        & 75.1                         \\ \hline
ResNet\_50v2    & 75.6                    & $75^{\small{*}}$                        & 75.4                         \\ \hline
\end{tabular}
\end{table}

\makeatletter
\newdimen\legendxshift
\newdimen\legendyshift
\newcount\legendlines
% distance of frame to legend lines
\newcommand{\bclldist}{1mm}
\newcommand{\bclegend}[3][7mm]{%
    % initialize
    \legendxshift=0pt\relax
    \legendyshift=0pt\relax
    \xdef\legendnodes{}%
    % get width of longest text and number of lines
    \foreach \lcolor/\ltext [count=\ll from 1] in {#3}%
        {\global\legendlines\ll\pgftext{\setbox0\hbox{\bcfontstyle\ltext}\ifdim\wd0>\legendxshift\global\legendxshift\wd0\fi}}%
    % calculate xshift for legend; \bcwidth: from bchart package; \bclldist: from node frame, inner sep=\bclldist (see below)
    % \@tempdima: half width of bar; 0.72em: inner sep from text nodes with some manual adjustment
    \@tempdima#1\@tempdima0.5\@tempdima
    \pgftext{\bcfontstyle\global\legendxshift\dimexpr\bcwidth-\legendxshift-\bclldist-\@tempdima-0.72em}
    % calculate yshift; 5mm: heigt of bar
    \legendyshift\dimexpr5mm+#2\relax
    \legendyshift\legendlines\legendyshift
    % \bcpos-2.5mm: from bchart package; \bclldist: from node frame, inner sep=\bclldist (see below)
    \global\legendyshift\dimexpr\bcpos-2.5mm+\bclldist+\legendyshift
    % draw the legend
    \begin{scope}[shift={(\legendxshift,\legendyshift)}]
    \coordinate (lp) at (0,0);
    \foreach \lcolor/\ltext [count=\ll from 1] in {#3}%
    {
        \node[anchor=north, minimum width=#1, minimum height=5mm,fill=\lcolor] (lb\ll) at (lp) {};
        \node[anchor=west] (l\ll) at (lb\ll.east) {\bcfontstyle\ltext};
        \coordinate (lp) at ($(lp)-(0,4mm+#2)$);
        \xdef\legendnodes{\legendnodes (lb\ll)(l\ll)}
    }
    % draw the frame
    \node[draw, inner sep=\bclldist,fit=\legendnodes] (frame) {};
    \end{scope}
}
\makeatother

\begin{figure}
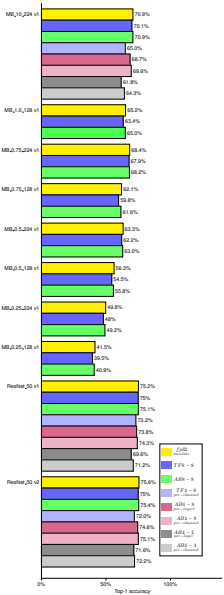

\scalebox{0.20}{
\centering
\begin{bchart}[step=50,max=140,unit=\%,scale=1.5]     
    \bcbar[label=MB\_10\_224 v1, color=yellow]{70.9}
    \bcbar[color=blue!60]{70.1}
    \bcbar[color=green!60]{70.9}
    \bcbar[color=blue!30]{65.0}
    \bcbar[color=purple!60]{68.7}
    \bcbar[color=purple!30]{69.6}
    \bcbar[color=gray!90]{61.8}
    \bcbar[color=gray!40]{64.3}

    \smallskip
    \bcbar[label=MB\_1.0\_128 v1, color=yellow]{65.2}
    \bcbar[color=blue!60]{63.4}
    \bcbar[color=green!70]{65.0}
%    \bcbar[color=purple!60]{62.0}

    \smallskip
    \bcbar[label=MB\_0.75\_224 v1, color=yellow]{68.4}
    \bcbar[color=blue!60]{67.9}
    \bcbar[color=green!60]{68.2}
%    \bcbar[color=purple!60]{65.0}
    
    \smallskip
    \bcbar[label=MB\_0.75\_128 v1, color=yellow]{62.1}
    \bcbar[color=blue!60]{59.8}
    \bcbar[color=green!60]{61.6}
%    \bcbar[color=purple!60]{58.2}

    \smallskip
    \bcbar[label=MB\_0.5\_224 v1, color=yellow]{63.3}
    \bcbar[color=blue!60]{62.2}
    \bcbar[color=green!60]{63.0}
%    \bcbar[color=purple!60]{58.4}

    \smallskip
    \bcbar[label=MB\_0.5\_128 v1, color=yellow]{56.3}
    \bcbar[color=blue!60]{54.5}
    \bcbar[color=green!60]{55.8}
%    \bcbar[color=purple!60]{50.5}

    \smallskip
    \bcbar[label=MB\_0.25\_224 v1, color=yellow]{49.8}
    \bcbar[color=blue!60]{48}
    \bcbar[color=green!60]{49.2}
%    \bcbar[color=purple!60]{43.8}
    
    \smallskip
    \bcbar[label=MB\_0.25\_128 v1, color=yellow]{41.5}
    \bcbar[color=blue!60]{39.5}
    \bcbar[color=green!60]{40.9}
%    \bcbar[color=purple!60]{35.5}

    \smallskip
    \bcbar[label=ResNet\_50 v1, color=yellow]{75.2}
    \bcbar[color=blue!60]{75}
    \bcbar[color=green!60]{75.1}
    \bcbar[color=blue!30]{73.2}
    \bcbar[color=purple!60]{73.8}
    \bcbar[color=purple!30]{74.3}
    \bcbar[color=gray!90]{69.6}
    \bcbar[color=gray!40]{71.2}

    \smallskip
    \bcbar[label=ResNet\_50 v2, color=yellow]{75.6}
    \bcbar[color=blue!60]{75}
    \bcbar[color=green!60]{75.4}
    \bcbar[color=blue!30]{72.0}
    \bcbar[color=purple!60]{74.6}
    \bcbar[color=purple!30]{75.1}
    \bcbar[color=gray!90]{71.6}
    \bcbar[color=gray!40]{72.2}
    
    \smallskip
    \bclegend{6pt}{
    yellow/$\underset{baseline}{fp32}$,
    blue!60/${TF 8-8}$,
    green!60/${AB 8-8}$,
    blue!30/$\underset{per-channel}{TF 4-8}$,
    purple!60/$\underset{per-layerl}{AB 4-8}$,
    purple!30/$\underset{per-channel}{AB 4-8}$,
    gray!90/$\underset{per-layer}{AB 4-4}$,
    gray!40/$\underset{per-channel}{AB 4-4}$
    }
    \bcxlabel{ Top-1 accuracy }
\end{bchart}}
\caption{Top-1 accuracy of fp32, Tensorflow(TF)'s and Alpha-blending(AB) optimization with 8-bits or 4-bits numerical forms. 8-8: 8-bits weight and activation; 4-8: 4-bits weight and 8-bits activation; 4-4: 4-bits weight and activation.}
\label{fig:cha_1}
\end{figure}

\subsubsection{Evaluation of 4-bits weight and 8-bits activation (INT4-8)} \label{sect:4.3}

\cite{TF-int4-8:} reported that accuracy of 4-bits weight and 8-bits activation (INT4-8) is within 5\% of the fp32 baseline for Mobilenet v1 and ResNet networks. We ran the same models using AB quantization, and have listed the result in the $4^{th}$ and $5^{th} $columns in table \ref{table2}. The $4^{th}$ one is per-layer quantization, and $5^{th}$ is per-channel. 

AB INT4-8 achieved a 1.53\% accuracy drop on average compared to the fp32 baseline for per-layer quantization, and a 0.9\% accuracy drop for per-channel quantization. Moreover, AB's INT4-8 per-channel performance outperforms the prior result ~\cite{TF-int4-8:} in $3^{rd}$ col. by 2.93\%.

%The compact models, e.g. MB\_0.25\_128 v1, MB\_0.25\_224 v 1, MB\_0.5\_128 v1, have wider accuracy degradation %(5-7\%) than the large size ones, e.g. MB\_1.0\_224 v1, ResNet\_50 v1,v2, which have 1-2\% accuracy loss only. 
%INT4-8 has 2X improvement on the weight memory size and 2X MAC power efficiency over the INT8-8. For large and accurate neural networks, this experiment justified that efficient machine learning hardware, i.e. CPU, GPU or accelerator, is in need of matrix multiply operation with mixed 4-bits and 8-bits operands.    

\begin{table}[h]
\caption{ Top-1 accuracy: the pre-trained accuracies are in $2^{nd}$ col.; Tensorflow's INT4-8 - 4bits weight and 8-bits activation are in $3^{rd}$ col.; $\textbf{AB}$ INT4-8 - 4-bits weight and 8-bits activation in $4^{th}$ and $5^{th}$ cols. Note: for MobileNet in table \ref{table2}, the first layer and all depth-wise convolution layers, which are only 1.1\% of all the weights and consume 5.3\% of total MAC operations for inference are quantized into 8-bits. For ResNet v1 and v2, weight and activation of the first layer are quantized into 8-bits. $^+$\protect\cite{TF-int4-8:} }

\label{table2}
\begin{tabular}{|l|c|c|c|c|}
\hline
Model name        & \multicolumn{1}{l|}{

{\begin{tabular}[c]{@{}l@{}}

{fp32} \\ {\ \ \%} \end{tabular}}

} & \multicolumn{1}{l|} {

{\begin{tabular}[c]{@{}l@{}}

\small{TF4-8} \\ {\ \ \ \%} \end{tabular}}

} & \multicolumn{1}{l|} {

{\begin{tabular}[c]{@{}l@{}}

\small{AB4-8} \\ {\scriptsize{per-layer\ \%}} \end{tabular}}

} &  \multicolumn{1}{l|} {

{\begin{tabular}[c]{@{}l@{}}

\small{AB4-8} \\ {\scriptsize{per-channel\ \%}} \end{tabular}}} \\ \hline

{\small MB1.0\_224v1}  & 70.9                    & $65.0^{\small{+}}$                        & 68.7                                                                                & 69.6                               \\ \hline
{\small MB0.75\_224v1}    & 68.4                    &   -                        & 65                                                                                &       -                                            \\ \hline
{\small MB0.50\_224v1}    & 63.3                    &  -                         & 58.4                                                                                &     -                                            \\ \hline
{\small MB0.25\_224v1}    & 49.8                    &   -                        & 43.8                                                                                &      -                                           \\ \hline
ResNet\_50v1    & 75.2                    & $73.2^{\small{+}}$
& 73.8                                                                                & 74.3                                                          \\ \hline
ResNet\_50v2    & 75.6                    & $72^{\small{+}}$                          & 74.6                                                                                & 75.1                                                          \\ \hline

\end{tabular}
\end{table}

\subsubsection{Evaluation of 4-bits weight and 4-bits activation (INT4-4)} \label{sect:4.4}

Finally, we quantized the well-known neural networks, MobileNet\_1.0\_224 v1 and ResNet\_50 v1/v2, using 4-bit weights and 4-bit activations. The $4^{th}$ column in table \ref{table-3} is for per-layer quantization, whose accuracy is 5.5\% lower than fp32's in average. The per-channel quantization in the $5^{th}$ column has 4.66\% accuracy loss. AB's INT4-4  result, using per-channel quantization, achieves similar accuracy as the TF4-8 scheme \cite{TF-int4-8:}, which has 4-bits weight and 8-bits activation as shown in the $3^{rd}$ column. 

% Given that a 4-bit MAC operation is about 50\% cost of operation with 4-bit and 8-bit mixed operands and consume 50\% memory size for activation, therefore this experiment probably demonstrated that the proposed Alpha-blending $\textbf{AB}$ is a more efficient quantization aware training flow. 

\begin{table}[h]
\caption{top-1 accuracy of fp32, Tensorflow's INT4-8 and $\textbf{AB} $ INT4-4 quantization. The first layers, all depth-wise layers and the last layer are quantized in to 8-bits, and all other layers are in 4-bits both for weight and activation. $^+$\protect\cite{TF-int4-8:}}
\label{table-3}
\begin{tabular}{|l|c|c|c|l|}
\hline

Model name       & \multicolumn{1}{l|}{

{\begin{tabular}[c]{@{}l@{}}
{fp32} \\ {\ \%} \end{tabular}}

} & \multicolumn{1}{l|}{

{\begin{tabular}[c]{@{}l@{}}
{\small{TF4-8}} \\ {\ \ \ \%}\end{tabular}}

} & \multicolumn{1}{l|}{\begin{tabular}[c]{@{}l@{}}
{\small{AB4-4}} \\ {\scriptsize{per-layer \%}} \end{tabular}} & \begin{tabular}[c]{@{}l@{}}{\small{AB4-4}}\\ {\scriptsize{per-channel \%}} \end{tabular} \\ \hline
{\small MB1.0\_224v1} & 70.9                    & $65.0^{\small{+}}$                        & 61.8                                                                                & 64.3                                                            \\ \hline
\small{ResNet\_50v1}   & 75.2                    & $73.2^{\small{+}}$                        & 69.6                                                                                & 71.2                                                          \\ \hline
\small{ResNet\_50v2}   & 75.6                    & $72^{\small{+}}$                          & 71.6                                                                                    & 72.2                                                              \\ \hline
\end{tabular}
\end{table}

\section{Conclusion and future work}

We have introduced alpha-blending ($\textbf{AB}$), an alternative method to the well-known $\textit{Straight-Through Estimator}$ ($\textbf{STE}$) for learning low precision neural networks using SGD. $\textbf{AB}$ accepts the almost everywhere zero gradient of quantization function during Backprop, and uses an affine combination of the original full-precision weights and corresponding quantized values as the actual weights in the loss function. This change allows the gradient update to the full-precision weights in backward propagation to be performed through the full-precision path incrementally, instead of applying STE to the quantization path.

To measure the impact on network accuracy using the AB methodology, we have trained a single-bit BinaryNet(BBN) \cite{binary} on CIFAR10 to show that $\textbf{AB}$ generates equivalent or better accuracy compared to training with $\textbf{STE}$. Moreover, we have applied the AB metholody to larger, more practical networks such as MobileNet and ResNet to compare with $\textbf{STE}$ based quantization. The top-1 accuracy of 8-bits weight and 8-bits activation is 0.82\% better than the existing state-of-art results \cite{TF-mb-int8:}\cite{TF-rn-int8:}. For 4-bits weight and 8-bits activation quantization, AB has 2.93\% higher top-1 accuracy on average compared to that reported in \cite{TF-int4-8:}.

$\textbf{AB}$ can also be applied to several other network optimization techniques besides quantization. We plan to investigate AB on clustering and pruning in a future work. 

%\section{Acknowledgment}

%% The file named.bst is a bibliography style file for BibTeX 0.99c
\bibliographystyle{named}
\bibliography{ijcai19}

\end{document}